\documentclass[11pt]{article}
\usepackage{acl2016}
\usepackage{times}
\usepackage{latexsym}
\usepackage{amssymb}
\usepackage{url}
\usepackage{graphicx}

\aclfinalcopy 
\def\aclpaperid{318} 

\usepackage{enumitem}
\setlist{nolistsep}

\title{Natural Language Generation enhances human decision-making with uncertain information}

\author{Dimitra Gkatzia\\
	    School of Computing\\
	    Edinburgh Napier University\\
	    Edinburgh, EH10 5DT, UK\\
	    {\tt d.gkatzia@napier.ac.uk}
	  \And
	Oliver Lemon\\
  	Interaction Lab\\
  	Heriot-Watt University\\
  	Edinburgh, EH14 4AS, UK\\
  {\tt o.lemon@hw.ac.uk}
    \And
	Verena Rieser\\
  	Interaction Lab\\
  	Heriot-Watt University\\
  	Edinburgh, EH14 4AS, UK\\
  {\tt v.t.rieser@hw.ac.uk}}

\date{}

\begin{document}

\maketitle

\begin{abstract}
Decision-making is often dependent on uncertain data, e.g.\ data associated with confidence scores or  probabilities.
We present a  comparison of different information presentations for uncertain data and, for the first time, measure their effects on human decision-making. 
We show that 
 the use of Natural Language Generation (NLG) improves decision-making under uncertainty,  
 compared to  state-of-the-art graphical-based representation methods. In a task-based study with 442 adults, we found that  presentations using NLG lead to 24\% better decision-making on average than the graphical presentations, and to 44\% better decision-making when NLG is combined with graphics.
We also show that 
women achieve significantly better results when   presented with NLG output (an  87\% increase on average compared to graphical presentations). 
\end{abstract}

\section{Introduction}

Natural Language Generation (NLG)  technology can achieve comparable results to commonly used data visualisation techniques for supporting accurate human decision-making \cite{Gatt2009}.
In this paper, we investigate whether NLG technology can also be used to support decision-making when the underlying data is uncertain. 
%
%
Current  data-to-text systems assume that the underlying data is precise and correct -- an assumption which is heavily criticised by other disciplines concerned with decision support, such as medicine \cite{Gigerenzer:11}, environmental modelling \cite{beven:2009}, climate change \cite{IPCC04}, or weather forecasting \cite{WMO:2008}.   
However, simply presenting  numerical expressions of risk and uncertainty is not enough.
Psychological studies on decision making have found that a high percentage of people do not understand and can't act upon numerical uncertainty \cite{Cokely:2012,galesic:2012}. For example, about 30\% of participants in a German-American study
  are unable to answer the question: ``{\em Which of the following numbers represents the biggest risk of getting a disease: 
1 in 100, 1 in 1000, 1 in 10?}'' \cite{galesic:2012}.



\begin{figure}[t]
    \centering
    \includegraphics[width=8cm]{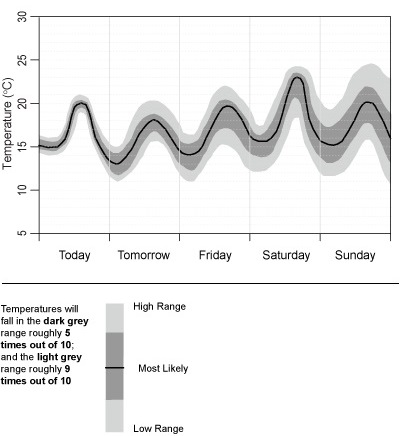}
    \caption{Graphics for temperature data.}
    \label{fig:screenshot:temp}
\end{figure}

So far, the NLG community has investigated the conversion of numbers into language \cite{power:12} and the use of vague expressions \cite{vanDeemter2009}.
In this work, we explore how to convert numerical representations of uncertainty into Natural Language so as to maximise confidence and correct outcomes of human decision-making. We consider the exemplar task of weather forecast generation. We initially present two NLG strategies which present the uncertainty in the input data. The two strategies are based on (1) the World Meteorological Organisation (WMO) \cite{WMO:2008} guidelines and (2) commercial forecast presentations (e.g.\ from BBC   presenters). We then evaluate the  strategies against a state-of-the-art graphical system \cite{Stephens2011}, which presents the uncertain data in a graphical way.  Figure \ref{fig:screenshot:temp} shows an example of this baseline graphical presentation.
We use a game-based setup 
  \cite{gkatzia-EtAl:2015:ENLG} to perform task-based evaluation,  to investigate the effect that the different information presentation strategies have on human decision-making.

Weather forecast generation is a common topic within the NLG community, e.g.\ 
 \cite{Konstas2012,Angeli2010,BelzKow2010,Sripada2005}. Previous approaches have not focused on how to communicate uncertain information or 
  the best ways of referring to probabilities of meteorological phenomena to occur. In addition, their evaluation is 
   based on user ratings of grammatically, semantic correctness, fluency, coherence or via post-edit evaluation. Although these metrics are indicative of the quality of the text produced, they do not measure the impact the texts might have in people's comprehension of uncertainty or on their ability to make decisions based on the information conveyed. 

Our contributions to the field are as follows: 
(1) We study a  principled mapping of uncertainty to Natural Language and provide recommendations and data for future NLG systems;
(2) We introduce a game-based data collection environment which extends task-based evaluation by measuring the impact of NLG on decision-making (measuring user confidence and game/task success); 
and (3) We show that effects of the different representations vary for different user groups, so that user adaptation is necessary when generating multi-modal presentations of uncertain information.


\begin{figure*}[t]
    \centering
    \includegraphics[width=12cm]{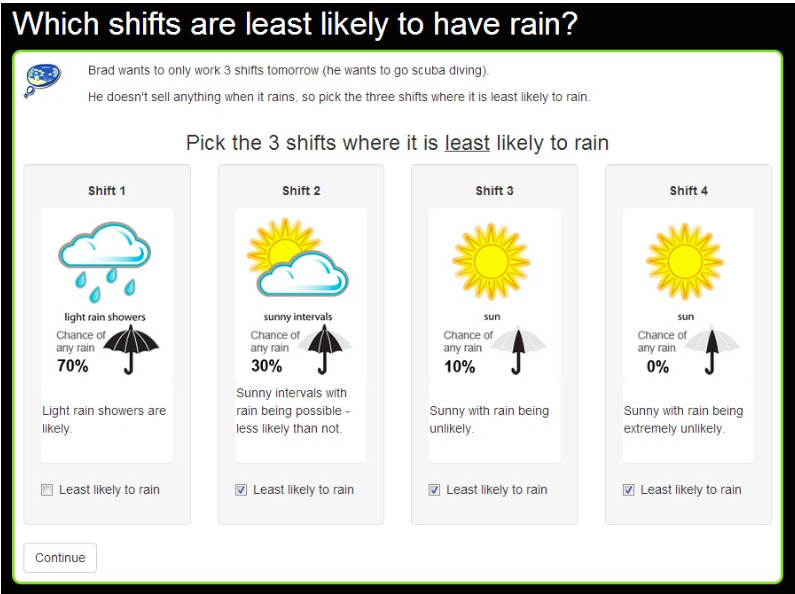}
    \caption{Screenshot of the Extended Weather Game (Rainfall: Graphics and  WMO condition).}
    \label{fig:screenshot}
\end{figure*}


\section{The Extended Weather Game} \label{weather-game}

In this section, we present our extended version of the MetOffice's Weather Game \cite{Stephens2011}. The player has to choose where to send an ice-cream vendor in order to maximise sales, given weather forecasts for four weeks and two locations. These forecasts describe (1) predicted rainfall (Figure \ref{fig:screenshot}) and (2)  temperature levels together with their likelihoods
 in three ways: (a) through graphical representations (which is the version of the original game), (b) through textual forecasts, and (c) through combined graphical and textual forecasts. We generated the textual format using two rule-based NLG approaches as described in the next section. Users are asked to initially choose the best destination for the ice-cream vendor 
 and then they are asked to state how confident they are with their choice. 
 Based on their decisions and their confidence levels, the participants are finally presented with their ``monetary gain". For example, the higher the likelihood of sunshine, the higher the monetary gain if the player has declared that s/he is confident that it is not going to rain and it doesn't actually rain. In the opposite scenario, the player would lose money. 
 The decision on whether rain occurred is estimated by sampling the probability distribution. 
 At the end of the game, users were scored according to their ``risk literacy" following the Berlin Numeracy Test \cite{Cokely:2012}. 
Further details are presented in \cite{gkatzia-EtAl:2015:ENLG}.

\section{Natural Language Generation from Uncertain Information} \label{NLG}

We developed two NLG systems, WMO-based and NATURAL, using SimpleNLG \cite{SimpleNLG}, which both generate textual descriptions of rainfall and temperature data addressing the uncertain nature of forecasts. 

\noindent{\bf WMO-based:} This is a rule-based system which uses the guidelines recommended by the WMO \cite{WMO:2008} for reporting uncertainty, as shown in Table \ref{probs}. 
Consider for instance a forecast of sunny intervals with 30\% probability of rain. This WMO-based system will generate the following forecast: ``Sunny intervals with rain being possible - less likely than not''.

\begin{table}
\small
\centering
\begin{tabular}{|p{2.5cm}|l|}
\hline \bf Likelihood of occurrence & \bf Lexicalisation \\ \hline \hline
p \textgreater 0.99 & ``extremely likely" \\
$0.90  \leq p \leq 0.99$  & ``very likely" \\
$0.70  \leq p \leq 0.89$  & ``likely" \\
$0.55  \leq p \leq 0.69$  & ``probable - more likely than not" \\
$0.45  \leq p \leq 0.54$  & ``equally likely as not" \\
$0.30  \leq p \leq 0.44$  & ``possible - less likely than not" \\
$0.10  \leq p \leq 0.29$  & ``unlikely" \\
$0.01 \leq p \leq 0.09$  & ``very unlikely" \\
$p \textless 0.01 $  & ``extremely unlikely" \\
\hline
\end{tabular}
\caption{\label{probs} WMO-based mapping of likelihoods.}
\end{table}

\noindent{\bf NATURAL:} This system imitates forecasters and their
natural way of reporting weather. The rules used in this system have been derived by observing the way that experts (e.g.\ BBC weather reporters) produce forecasts. For the previous example (sunny intervals with 30\% probability of rain), this system will generate the following forecast: ``Mainly dry with sunny spells''.

\section{Evaluation} \label{evaluation}

In order to investigate what helps people to  better understand and act upon uncertainty in information presentations, we use five conditions within the context of the Extended Weather Game:
\begin{enumerate}
\item \textbf{Graphics only:} This representation shows the users only the graphical representation of the weather forecasts. For this condition we used the graphs that scored best in terms of human comprehension from \cite{Stephens2011}.
\item \textbf{Multi-modal Representations:}\\
{$-$ \bf Graphics and NATURAL:} This is a multi-modal representation consisting of graphics (as described in the previous condition) and text produced by the NATURAL system.\\
{$-$ \bf Graphics and WMO-based:} This is also a multi-modal representation consisting of graphics and text produced by the WMO-based system.
\item \textbf{NLG only:}\\
{$-$ \bf NATURAL only:} This is a text-only representation as described above. \\
{$-$ \bf WMO-based system only:} This is also a text-only representation. 
\end{enumerate}

\section{Data} \label{sec:data}
We recruited 442 unique players (197 females\footnote{Women made up 44.5\% of the subjects.}, 241 males, 4 non-disclosed) using social media. We collected 450 unique game instances (just a few people played the game twice). 
The anonymised data will be released as part of this submission.

\section{Results} \label{results}


In order to investigate which representations assist people in decision-making under uncertainty, we analysed both the players' scores (in terms of monetary gain) and their predictions for rainfall with regard to their confidence scores.
 As we described in Section \ref{weather-game}, the game calculates a monetary gain based on both the decisions and the confidence of the player, i.e.\ the decision-making ability of the player.
Regarding confidence,  we asked users to declare how confident they are on a 10-point scale. 
 In our analysis 
 we therefore focus on both confidence and score at the game.



\subsection{Results for all adults} 

{\bf Multi-modal vs.\ Graphics-only:}
We found that use of  multi-modal representations  leads to gaining significantly higher game scores (i.e.\ better decision-making) than the Graphics-only representation ($p = 0.03$, effect = +$36.36$). 
 This is a 44\% average increase in game score.\\
{\bf Multi-modal vs.\ NLG-only:}
However, there is no significant difference between the NLG only and the multi-modal representation, for game score.\\
{\bf NLG vs.\ Graphics-only:}
We found that the NLG representations resulted in a 24.8\% increase in average task score (i.e.\ better decision-making) compared to the Graphics-only condition, see Table \ref{tab:results}: an average   score increase of over 20 points.
 There was no significant difference found between the WMO and NATURAL NLG conditions.\\
\noindent
{\bf Confidence:} For confidence, the multi-modal representation is significantly more effective than NLG only ($p<0.01$, effect = $17.7\%$). However, as Table \ref{tab:results} shows, although adults did not feel very confident when presented with NLG only, they were able to make better decisions compared to being presented with graphics only.

\noindent
{\bf Demographic factors:}
We further found that prior experience on making decisions based on risk, familiarity with weather models, and correct literacy test results are predictors of the players' understanding of uncertainty, which is translated in both confidence and game scores. In contrast, we found that the education level, the gender, or being native speaker of English does not contribute to players' confidence and game scores.

\begin{table}
\small
\centering
\begin{tabular}{|p{2.5cm}|l|p{2cm}|}
\hline   & \bf Monetary gains & \bf Confidence  \\ \hline \hline
Graphs only & 81.15 & 78.5\% \\
Multi-modal & 117.51& 83.7\%\\
NLG only &101.33 & 66\%\\
\hline
\end{tabular}
\caption{\label{tab:results} Average Monetary gains and Confidence scores (All Adults).}
\end{table}

\subsection{Results for Females} We found that females score significantly higher at the decision task  when exposed to either of the NLG output presentations, when compared to the graphics-only presentation ($p < 0.05$, effect = +$53.03$). This is an increase  of 87\%, also see Table \ref{tab:female}. 
In addition, the same group of users scores significantly higher when presented with the multi-modal output as compared to graphics only ($p = 0.05$, effect =$ 60.74\%$). Interestingly, for this group, the multi-modal presentation adds little more in effectiveness of decision-making than the NLG-only condition, but the multi-modal presentations do enhance their confidence (+15\%).
We furthermore found that educated (i.e.\ holding a BSc or higher degree) females, who also correctly answered the risk literacy test, feel significantly more confident  when presented with the multi-modal representations than with NLG only ($p = 0.01$, effect = $16.7\%$). 

\begin{table}
\small
\centering
\begin{tabular}{|p{2.5cm}|l|p{2cm}|}
\hline   & \bf Monetary gains & \bf Confidence  \\ \hline \hline
Graphs only & 60.83 & 74.6\% \\
Multi-modal & 118.41& 81.3\%\\
NLG only &113.86 & 65.8\%\\
\hline
\end{tabular}
\caption{\label{tab:female} Average Monetary gains and Confidence scores (Females).}
\end{table}

\subsection{Results for Males} We found that males obtained similar game scores
with all the types of representation. 
This suggests that the overall improved scores (for All Adults) presented above, are largely due to the beneficial  effects of NLG for women. 
In terms of confidence, males are more likely to be more confident if they are presented with graphics only (81\% of the time) or a multi-modal representation (85\% of the time) ($p = 0.01$).










\section{Conclusions and Future Work} \label{conclusions}

We present results from a game-based study on how to generate descriptions of uncertain data -- an issue which so far has been unexplored by data-to-text systems.
We find that there are significant gender differences between multi-modal, NLG, and graphical  versions of the task, where for women, use of NLG results in a 87\% increase in task success over graphics. Multimodal presentations lead to a 44\% increase for all adults, compared to   graphics. People are also more confident of their judgements when using the multimodal representations.
%
These are significant findings, as previous work has not distinguished between genders when comparing different representations of data, e.g. \cite{Gatt2009}.
It also confirms 
research on gender effects in multi-modal systems, as for example reported in \cite{mef:eacl06,rieser:acl08,moller:2012}. 
The results are also related to educational research, which shows that women perform better in verbal-logical tasks than visual-spatial tasks \cite{Zhu2007}.
An interesting investigation for future research is the interplay between uncertainty, risk-taking behaviour and gender, as for example reported in \cite{Sarin20161}. 

\footnotesize
\section*{Acknowledgments}
\vspace{-0.2cm}
This research received funding from the EPSRC projects GUI (EP/L026775/1), DILiGENt (EP/M005429/1) and  MaDrIgAL (EP/N017536/1).
\normalsize

\bibliography{acl2016}
\bibliographystyle{acl2016}

\end{document}